\documentclass[runningheads]{llncs}

\usepackage{eccv}

\usepackage{eccvabbrv}

\usepackage{graphicx}
\usepackage{booktabs}
\usepackage{marvosym}
\usepackage[accsupp]{axessibility}  

\usepackage{hyperref}

\usepackage{orcidlink}
\usepackage{float}

\begin{document}

\title{\raisebox{-0.18cm}{\includegraphics[scale=0.39]{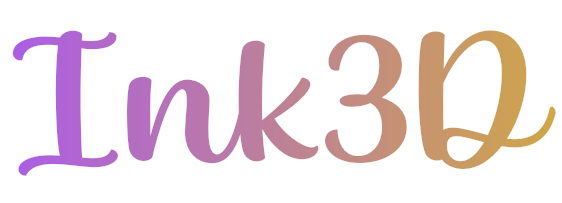}} Sculpting 3D Assets with Extremely Complex Textures via Video Generative Models} 


\titlerunning{Ink3D}

\makeatletter
\newcommand{\printfnsymbol}[1]{%
  \textsuperscript{\@fnsymbol{#1}}%
}

\makeatother

\author{Yue Han\inst{1,3}\texorpdfstring{\thanks{Equal contribution.}}{}, Chong Li\inst{2}\printfnsymbol{1}, Zhening Liu\inst{4}, Cong Huang\inst{1}, Fang Deng\inst{1}, Yong Liu\inst{3}, \\
Fangyun Wei\inst{2}\textsuperscript{\Letter}, Yan Lu\inst{2}
}

\authorrunning{Y. Han et al.}

\newcommand{\ghlink}{https://github.com/YueHan99/Ink3D.TextureGen}
\newcommand{\githubicon}{\raisebox{-1.5pt}{\includegraphics[height=1.05em]{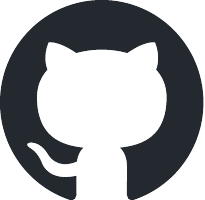}}}

\institute{
  \textsuperscript{1}ZGCA \& ZGCI \quad
  \textsuperscript{2}Microsoft Research \quad
  \textsuperscript{3}Zhejiang University \quad
  \textsuperscript{4}HKUST \quad \\
  {\small \texttt{12432015@zju.edu.cn} \quad \texttt{\{chol, fawe, yanlu\}@microsoft.com}}\\
  {\small \githubicon\ \url{https://github.com/YueHan99/Ink3D.TextureGen}}
}

\maketitle

\newcommand{\nonumberfootnote}[1]{%
  \renewcommand{\thefootnote}{}
  \footnotetext{\hspace*{-0.9em}#1}
  \renewcommand{\thefootnote}{\arabic{footnote}}
}

\nonumberfootnote{\textsuperscript{\Letter}\ Corresponding author.}

\begin{center}
     \centering
     \captionsetup{type=figure}
     \includegraphics[width=\textwidth]{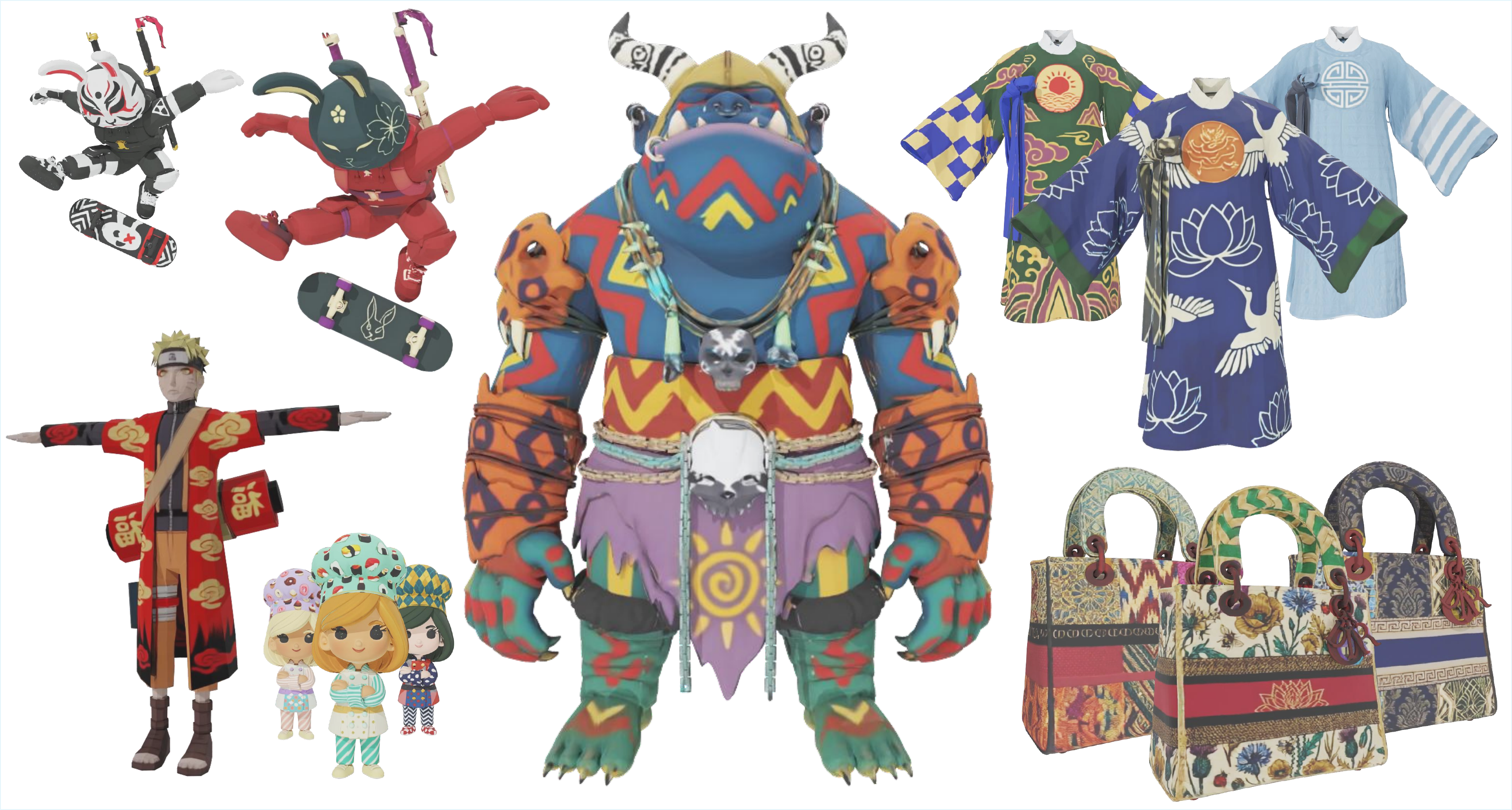}
     \vspace{-4mm}
    \captionof{figure}{\textit{Ink3D} introduces geometry-aware video generation into the 3D generation pipeline to synthesize 3D assets with extremely complex textures such as intricate clothing patterns and high-frequency decorative details.}
    \label{fig:teaser}
 \end{center}

\begin{abstract}
Recent 3D generative models can synthesize high-quality geometry but often struggle to reproduce intricate textures from reference images, largely due to the scarcity of large-scale 3D training data with rich surface appearance. In contrast, visual generative models are trained on datasets several orders of magnitude larger and excel at modeling complex visual patterns. Motivated by this gap, we introduce Ink3D, a framework that bridges 3D generation with large-scale video generative models to synthesize \textit{extremely complex textures}. Ink3D first reconstructs a white-mesh geometry using an off-the-shelf 3D generation model. It then employs OrbitPainter, a conditional video generative model, to produce dense orbit-scan videos capturing object appearance across viewpoints. To convert these views into coherent textures, we introduce TextureOptimizer, a neural baking module that integrates dense multi-view observations while mitigating geometry inconsistencies arising from video generation. By decoupling geometry and texture synthesis and leveraging large-scale pretrained video priors, Ink3D enables significantly richer and more faithful texture generation than prior approaches.
  \keywords{Texture generation \and Video generative models}
\end{abstract}  
\section{Introduction}

Recent years have witnessed remarkable progress in 3D generation~\cite{deitke2023objaverse, deitke2024objaversexl, zhang2025texverse, xiang2025trellis1, xiang2025trellis2, zhao2025hunyuan3d, hunyuan3d2025, li2025triposg, huang2025mv, yuan2025seqtex, zhang2024clay, yang2025pandora3d, li2025step1x, chen20243dtopia, chen2025dora, wu2025direct3ds2, he2025sparseflex}. 
Inspired by the success of 2D image synthesis~\cite{wu2025qwen,rombach2022high,cai2025z}, mainstream approaches~\cite{xiang2025trellis1,xiang2025trellis2,zhao2025hunyuan3d,hunyuan3d2025} largely adopt the latent diffusion paradigm: a 3D asset is first encoded into a compact latent space using a variational autoencoder, after which a diffusion model~\cite{xiang2025trellis1, xiang2025trellis2, chen2025dora, wu2025direct3ds2, he2025sparseflex} is trained to model the distribution in this latent space, enabling high-quality 3D synthesis. 
This design has substantially advanced geometric modeling, producing plausible shapes with increasingly refined structural details.

However, despite impressive geometric quality, existing 3D generation models often struggle to synthesize \emph{extremely complex textures}. 
Typical failure cases include text-rich surfaces (e.g., logos, signage, printed words), intricate clothing patterns (e.g., plaid fabrics, embroidery, layered motifs), or high-frequency decorative details. 
The underlying reason is straightforward: high-quality 3D training data are scarce and expensive to construct. 
In contrast, large-scale image and video data on the Internet exceed 3D datasets by several orders of magnitude, often reaching billions of samples. 
Consequently, image and video generative models~\cite{wan2025wan,wu2025qwen,cai2025z} pretrained on such massive corpora demonstrate superior visual modeling capability and remarkable generalization, especially for fine-grained appearance and texture-rich content.

These observations inspire us to \emph{decouple geometry and texture modeling}.
Instead of forcing a single 3D generative model to learn both structure and appearance from limited 3D data, we first generate a high-quality \emph{white-mesh} geometry using an existing 3D generation model~\cite{xiang2025trellis2}, and then leverage powerful visual generative models~\cite{wan2025wan} to synthesize texture information. 
Finally, the generated texture is refined and baked onto the underlying geometry, producing a fully textured 3D asset.

In this work, we introduce \textbf{Ink3D}, a framework that integrates 3D generation with large-scale video generative models. Ink3D can synthesize extremely complex textures, as illustrated in Figure~\ref{fig:teaser}. Given a reference image, Ink3D first employs an off-the-shelf 3D generation model~\cite{xiang2025trellis2} to reconstruct the corresponding white-mesh geometry. 
Texture synthesis is then performed through a dedicated two-stage pipeline:

\emph{OrbitPainter.} 
OrbitPainter is a conditional video generative model designed to produce dense multi-view observations of the object. 
Conditioned on the reference image, it generates two \textit{360-degree orbit-scan videos}: one with horizontal camera rotation and one with vertical rotation. 
Together, these two trajectories ensure comprehensive viewpoint coverage of the object surface. 
Benefiting from large-scale pretraining on massive video data, the model possesses strong visual generalization capability, requiring only lightweight fine-tuning to adapt to arbitrary object categories.

\emph{TextureOptimizer.}
The generated orbit-scan videos provide a dense set of multi-view images that can be used for texture baking. A straightforward approach is to directly consolidate these views and bake them onto the geometry. In practice, however, this naive strategy often produces noticeable artifacts, such as color bleeding and blurred textures. The main reason is that the generated views may exhibit geometry inconsistencies introduced during the video generation process. To address this issue, we introduce \emph{TextureOptimizer}, a neural baking framework that robustly integrates the dense-view observations while enforcing geometric consistency. Specifically, TextureOptimizer first converts the dense-view image set into a textured voxel representation guided by the underlying geometry, and then refines the voxelized texture using a prior-guided denoising process. This strategy effectively leverages the rich appearance information from multi-view observations while mitigating the geometry inconsistencies introduced during video generation.

Decomposing 3D generation into geometry-first and texture-second has been explored in prior works. 
Most existing methods~\cite{richardson2023texture,liu2023syncdreamer,tang2023mvdiffusion,zeng2024paint3d,liu2024syncmvd,shaikh2024paintit,zhao2025hunyuan3d, hunyuan3d2025,li2025step1x,zhang2024clay} rely on multi-view image generation, typically synthesizing six discrete views conditioned on a reference image and a given geometry. 
However, such sparse-view strategies suffer from two major limitations. 
First, due to incomplete view coverage, certain surface regions may remain untextured, leading to missing or blurred areas. 
Second, independently generated views often exhibit cross-view inconsistencies, where the same surface region appears with different colors or patterns across views. 
In contrast, our use of video generative models effectively alleviates both issues. 
The generated 360-degree orbit videos provide dense viewpoint coverage, ensuring full surface observation. 
Moreover, the intrinsic spatiotemporal consistency of video generation enforces coherent appearance across adjacent frames, naturally translating into strong cross-view consistency in 3D space.

\vspace{-2mm}
\section{Related Work}
\vspace{-2mm}
\noindent\textbf{3D Generation.}
Driven by the expansion of large-scale 3D datasets \cite{deitke2023objaverse, deitke2024objaversexl, zhang2025texverse}, 3D asset generation has made remarkable progress. Recently, several series of top-tier foundation models have achieved impressive results. Notable examples include the Trellis family \cite{xiang2025trellis1, xiang2025trellis2}, the Hunyuan3D series \cite{zhao2025hunyuan3d, hunyuan3d2025}, and the VAST generation models \cite{li2025triposg, huang2025mv, yuan2025seqtex}. Generally, these methods follow one of two main paths. The first is native 3D generation, which produces both geometry and texture simultaneously in a single model (e.g., the Trellis series~\cite{xiang2025trellis1, xiang2025trellis2}). The second is a decoupled pipeline that first generates the 3D geometry and then synthesizes the texture, as adopted by methods such as the Hunyuan3D series~\cite{zhao2025hunyuan3d, hunyuan3d2025}, models from the VAST family~\cite{li2025triposg, huang2025mv, yuan2025seqtex}, Clay \cite{zhang2024clay}, Pandora3D \cite{yang2025pandora3d}, Step1X-3D \cite{li2025step1x}, and 3DTopia-XL \cite{chen20243dtopia}. Regardless of the chosen path, generating high-quality 3D shapes has become highly successful. With recent advances in 3D VAEs and shape generation models, such as DoRA \cite{chen2025dora}, Direct3D-s2 \cite{wu2025direct3ds2}, SparseFlex \cite{he2025sparseflex}, and the geometry module of Trellis-2, obtaining robust geometric structures has become significantly easier. However, synthesizing extremely complex textures (e.g., text-rich patterns or intricate clothing fabrics) remains a major bottleneck. This limitation largely stems from the severe scarcity of high-quality 3D texture data.

\noindent\textbf{Geometry Texturing.}
Early attempts at texturing 3D geometry leverage the strong visual priors of 2D image diffusion models. Methods such as DreamFusion \cite{poole2022dreamfusion}, Magic3D \cite{lin2023magic3d}, ProlificDreamer \cite{wang2024prolificdreamer}, and Fantasia3D \cite{chen2023fantasia3d} employ Score Distillation Sampling (SDS) to optimize rendered images so that they align with the 3D geometry. However, these optimization-based approaches often produce over-smoothed textures and baked-in lighting artifacts. To improve texture fidelity and reduce such artifacts, subsequent works explore directly generating UV textures by unwrapping the 3D mesh into a 2D plane. Representative methods include Text2Tex \cite{chen2023text2tex} and TexGen \cite{yu2024texgen}. While this strategy allows texture synthesis to be performed in the UV space, the unwrapping process frequently introduces stretching distortions and noticeable seams. Recently, multi-view generation followed by re-projection has become the dominant paradigm for geometry texturing. These methods~\cite{richardson2023texture,liu2023syncdreamer,tang2023mvdiffusion,zeng2024paint3d,liu2024syncmvd,shaikh2024paintit} first generate a sparse set of discrete views (e.g., four or six images) and then project them back onto the 3D surface. Moreover, the texturing modules of large 3D systems, such as Hunyuan3D-Paint \cite{zhao2025hunyuan3d, hunyuan3d2025}, Step1X-3D \cite{li2025step1x}, and Clay \cite{zhang2024clay}, also rely heavily on this sparse multi-view strategy. However, sparse-view approaches suffer from two fundamental limitations. First, as pointed out by UniTEX \cite{liang2025unitex}, a limited number of views cannot fully cover the entire object surface, inevitably leaving occluded regions. Second, independently generated views often lack cross-view consistency, resulting in visible seams after projection. These limitations highlight the importance of dense and continuous viewpoint coverage for achieving high cross-view consistency. While most geometry texturing methods rely on image generation models, SeqTex \cite{yuan2025seqtex} attempts to leverage a video generative model. However, it still produces only a sparse set of views and therefore inherits the fundamental limitations of sparse viewpoint coverage.

\noindent\textbf{Video Generative Models.}
The field of video generation has advanced rapidly in recent years, evolving from early UNet-based latent diffusion models~\cite{blattmann2023stable, chen2023videocrafter1, chen2024videocrafter2, zeng2024make} to large-scale Diffusion Transformers~\cite{peebles2023scalable, esser2024scaling}. This architectural shift has enabled a new generation of powerful video foundation models, including open-source systems~\cite{yang2024cogvideox, ma2025step, kong2024hunyuanvideo, wan2025wan, team2025longcat} as well as high-performance proprietary models~\cite{sora, veo, Kling}. These models exhibit improved modality alignment across text, image, and video, enabling a range of foundational capabilities such as text-to-video generation, image-to-video synthesis, controllable generation~\cite{hu2025hunyuancustom,chen2025omniinsert,luo2025camclonemaster,bai2025recammaster,zhao2025spatia}, and world modeling~\cite{agarwal2025cosmos,bruce2024genie,alonso2024diffusion}. As a result, they have also unlocked a wide range of applications in domains such as embodied AI~\cite{shen2025videovla,zhou2024robodreamer,zhang2025gevrm}, gaming~\cite{li2025hunyuan,wang2025animate,feng2024matrix}, and virtual avatar generation~\cite{gao2025wan,gan2025omniavatar,tu2025stableavatar}. 
\section{Ink3D}


\noindent\textbf{Problem Formulation.}
We adopt a geometry-first, texture-second paradigm for 3D asset generation. 
Given a reference image $\boldsymbol{I}$, we first employ an off-the-shelf 3D reconstruction model~\cite{xiang2025trellis2} to estimate a white-mesh geometry $\mathcal{G}$, which captures the object’s structural shape without texture. 
The objective of Ink3D is to generate a high-quality texture consistent with $\boldsymbol{I}$ and map it onto $\mathcal{G}$, resulting in a fully textured 3D asset.

\begin{figure*}[!t]
    \centering
    \includegraphics[width=\linewidth]{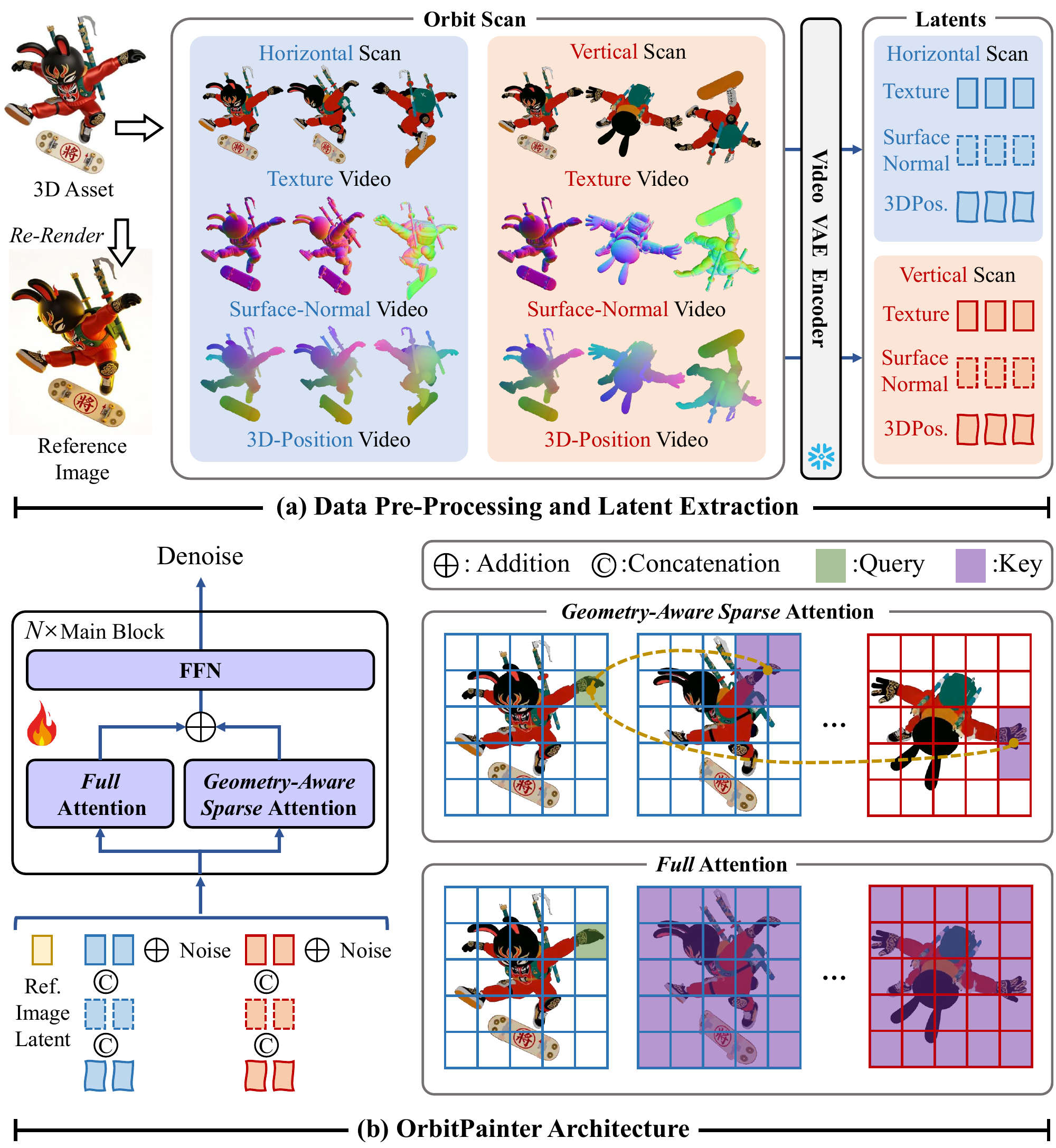}
    \vspace{-4mm}
    \caption{Overview of \textit{OrbitPainter}. (a) Given a 3D asset, we first apply random lighting and render it from a randomly sampled viewpoint to obtain a reference image. We then perform horizontal and vertical orbit scans to render RGB videos, surface-normal videos, and 3D-position videos. All videos, together with the reference image, are encoded by a video VAE encoder into latent representations. (b) OrbitPainter is a conditional video generative model that synthesizes RGB orbit-scan videos conditioned on a 2D appearance prior (the reference image) and 3D geometry priors (surface-normal and 3D-position videos). A Geometry-Aware Sparse Attention module is introduced in parallel with full attention to better leverage the 3D geometry priors.}
    \vspace{-4mm}
    \label{fig:training}
\end{figure*}

\noindent\textbf{Overview.}
Ink3D consists of two key components:

\begin{enumerate}
    \item \textit{OrbitPainter} (Section~\ref{sec:OrbitPainter}). OrbitPainter (Figure~\ref{fig:training}) is a conditional video generative model that incorporates a newly proposed Geometry-Aware Sparse Attention module (Section~\ref{sec:sparseattention}). Given the reference image $\boldsymbol{I}$ and the white-mesh geometry $\mathcal{G}$, it synthesizes two orbit-scan videos of the object: a horizontal orbit video $\boldsymbol{V}_{H}$ and a vertical orbit video $\boldsymbol{V}_{V}$. These videos ensure comprehensive viewpoint coverage while maintaining appearance consistency with $\boldsymbol{I}$, and geometry consistency with $\mathcal{G}$.
    \item \textit{TextureOptimizer} (Section~\ref{sec:TextureOptimizer}). 
TextureOptimizer (Figure~\ref{fig:inference}) consolidates frames from the generated orbit videos to bake textures onto $\mathcal{G}$. It explicitly mitigates geometric inconsistencies arising from imperfect video generation and resolves the challenges of dense multi-view texture fusion, resulting in a coherent and high-fidelity texture map.
\end{enumerate}

\vspace{-3mm}
\subsection{OrbitPainter}
\label{sec:OrbitPainter}

\noindent\textbf{Training Data Pre-Processing.}
As illustrated in Figure~\ref{fig:training}(a), given a textured 3D asset from the training set, we first randomly sample a viewpoint and apply random lighting, then render a static image as the reference image $\boldsymbol{I}$. We subsequently predefine two camera orbit trajectories: a horizontal scan and a vertical scan. For each trajectory, we render multiple modalities to construct supervision signals.

\begin{itemize}
    \item \textit{Horizontal Scan.}
Along the horizontal orbit trajectory, we render: (1) an RGB video $\boldsymbol{V}^{RGB}_H$,  
(2) a surface-normal video $\boldsymbol{V}^{Normal}_H$, and  
(3) a 3D-position video $\boldsymbol{V}^{Pos}_H$. Note that each video consists of 61 frames by default and is rendered at a resolution of $512\times 512$.
\item \textit{Vertical Scan.}
Similarly, along the vertical orbit trajectory, we render the corresponding RGB, surface-normal, and 3D-position videos: $\boldsymbol{V}^{RGB}_V$, $
\boldsymbol{V}^{Normal}_V$, and 
$\boldsymbol{V}^{Pos}_V$.
\end{itemize}

The horizontal and vertical scans provide complementary viewpoint coverage and multi-modal geometric supervision for training.

\noindent\textbf{Latent Extraction.}
Since we adopt a latent video diffusion model as the backbone, we first encode all rendered videos into latent representations. 
Specifically, we extract latents from 
\{$\boldsymbol{V}^{RGB}_H$, $\boldsymbol{V}^{Normal}_H$, $\boldsymbol{V}^{Pos}_H$, 
$\boldsymbol{V}^{RGB}_V$, $\boldsymbol{V}^{Normal}_V$, $\boldsymbol{V}^{Pos}_V$\}. We employ the WAN 2.2~\cite{wan2025wan} video VAE encoder, which uses a spatial downsampling rate of 16 and a temporal downsampling rate of 4, to obtain compact spatiotemporal latent features. 
The resulting latent representations are denoted as 
\{$\boldsymbol{F}^{RGB}_H$, $\boldsymbol{F}^{Normal}_H$, $\boldsymbol{F}^{Pos}_H$, 
$\boldsymbol{F}^{RGB}_V$, $\boldsymbol{F}^{Normal}_V$, $\boldsymbol{F}^{Pos}_V$\}. In addition, since the video encoder also supports image encoding, we use the same encoder to map the reference image $\boldsymbol{I}$ into its latent representation $\boldsymbol{F}_{I}$.

\noindent\textbf{Architecture.} Figure~\ref{fig:training}(b) illustrates the architecture of OrbitPainter, a conditional video generative model. 
The model is conditioned on two types of priors: 
\begin{enumerate}
    \item A 2D appearance prior $\boldsymbol{F}_{I}$, extracted from the reference image.
    \item 3D geometric priors 
$\{\boldsymbol{F}^{Normal}_H, \boldsymbol{F}^{Pos}_H, 
\boldsymbol{F}^{Normal}_V, \boldsymbol{F}^{Pos}_V\}$, 
which encode surface orientation and spatial structure under horizontal and vertical scans.
\end{enumerate}

Given these conditional inputs, the training objective is to generate the RGB latent videos 
$\{\boldsymbol{F}^{RGB}_H, \boldsymbol{F}^{RGB}_V\}$ using a Flow Matching objective. 
The generated latents represent multi-view appearance in video form while remaining consistent with the underlying 3D geometry.

\paragraph{Flow Construction.}
We first concatenate the target RGB latent videos along the temporal dimension:
\begin{equation}
\boldsymbol{F}^{RGB} = \text{Concat}_{T}(\boldsymbol{F}^{RGB}_H, \boldsymbol{F}^{RGB}_V),
\end{equation}
where $\text{Concat}_T(\cdot)$ denotes temporal concatenation. 
Following the Flow Matching formulation, we sample $t \in [0,1]$ from a logit-normal distribution and initialize Gaussian noise $\mathbf{x}_0 \sim \mathcal{N}(\mathbf{0},\mathbf{I})$. 
The intermediate sample is obtained via linear interpolation:
\begin{equation}
\mathbf{x}_t = (1-t)\mathbf{x}_0 + t\boldsymbol{F}^{RGB}.
\end{equation}

\paragraph{Injecting Geometric Priors.}
To guide the denoising process with geometry information, we first concatenate the horizontal and vertical geometric latents:
\begin{align}
\boldsymbol{F}^{Normal} &= \text{Concat}_T(\boldsymbol{F}^{Normal}_H,\boldsymbol{F}^{Normal}_V), \\
\boldsymbol{F}^{Pos} &= \text{Concat}_T(\boldsymbol{F}^{Pos}_H,\boldsymbol{F}^{Pos}_V).
\end{align}
These geometric priors are then injected by channel-wise concatenation with the intermediate latent:
\begin{equation}
\mathbf{x}'_t = \text{Concat}_C(\mathbf{x}_t,\boldsymbol{F}^{Pos},\boldsymbol{F}^{Normal}),
\end{equation}
where $\text{Concat}_C(\cdot)$ denotes channel-wise concatenation.

\paragraph{Injecting Appearance Prior.}
The 2D appearance prior is incorporated by temporally appending the reference image latent:
\begin{equation}
\boldsymbol{S} = \text{Concat}_T(\boldsymbol{F}_{I},\mathbf{x}'_t).
\end{equation}
The resulting sequence $\boldsymbol{S}$ therefore contains noise information together with both 2D appearance and 3D geometric priors.

\paragraph{Network Architecture.}
The sequence $\boldsymbol{S}$ is fed into the OrbitPainter network. 
As shown in Figure~\ref{fig:training}(b), OrbitPainter consists of $N$ Transformer blocks. 
Each block contains two parallel attention modules: a standard full-attention module and a proposed \textit{Geometry-Aware Sparse Attention} module (introduced in Section~\ref{sec:sparseattention}). 
The full-attention module processes the entire sequence $\boldsymbol{S}  = \text{Concat}_T(\boldsymbol{F}_{I},\mathbf{x}'_t)$, while the sparse attention module operates only on the video-related tokens $\mathbf{x}'_t$ to better capture geometry-aware correspondences.

\paragraph{Training Objective.}
We adopt Flow Matching as the training objective. 
The model learns to predict the velocity $\mathbf{u}_t = d\mathbf{x}_t/dt$ by minimizing the mean squared error between the predicted velocity $\mathbf{v}_t$ and the ground-truth velocity:
\begin{equation}
\label{eq:loss}
\mathcal{L} =
\mathbb{E}_{t,\mathbf{x}_0,\boldsymbol{F}^{RGB}}
\left\|
\mathbf{v}_t - \mathbf{u}_t
\right\|^2 .
\end{equation}

\subsection{Geometry-Aware Sparse Attention}
\vspace{-1mm}
\label{sec:sparseattention}

We formulate 3D texture generation as a video generation problem, where the goal is to synthesize horizontal and vertical orbit-scan videos conditioned on a reference image prior and 3D geometry priors. This setting differs from conventional video generation tasks, which typically rely only on appearance priors (e.g., a reference image) without explicit geometric information. The availability of 3D geometry priors therefore provides an opportunity to guide the generation process more effectively.

In standard video diffusion models, the attention module performs full spatiotemporal attention. As illustrated in Figure~\ref{fig:training}(b), each query token attends to all tokens across both spatial and temporal dimensions. This design is necessary because traditional video generation lacks structural priors that could constrain the attention space.

In our setting, the video frames are generated from orbit scans of a 3D asset, and explicit geometric priors are available. This enables us to leverage the underlying 3D geometry to guide the attention mechanism. Specifically, each spatiotemporal token in the latent video corresponds to a surface point on the 3D mesh. Given a query token, we can recover its corresponding 3D location through the position-map representation. Based on this 3D correspondence, we restrict the attention to tokens that correspond to nearby or identical surface locations across different frames, rather than attending to all tokens in the video, as illustrated in Figure~\ref{fig:training}(b).

We refer to this module as \textit{Geometry-Aware Sparse Attention}. 
In practice, given a query spatiotemporal token $\mathcal{T}$, we first re-project it onto the underlying 3D geometry using its associated 3D position, thereby identifying the corresponding voxel location $(i,j,k)$ in the 3D space. 
We then define a local neighborhood around $(i,j,k)$ with an expansion radius $r$ (set to 3 by default).  
For tokens in other video frames, we select those whose associated 3D positions fall within this neighborhood. 
Only these tokens are used as keys in the attention computation for the query token $\mathcal{T}$, while all other tokens are ignored.

\begin{figure*}[htbp]
    \centering
    \includegraphics[width=\linewidth]{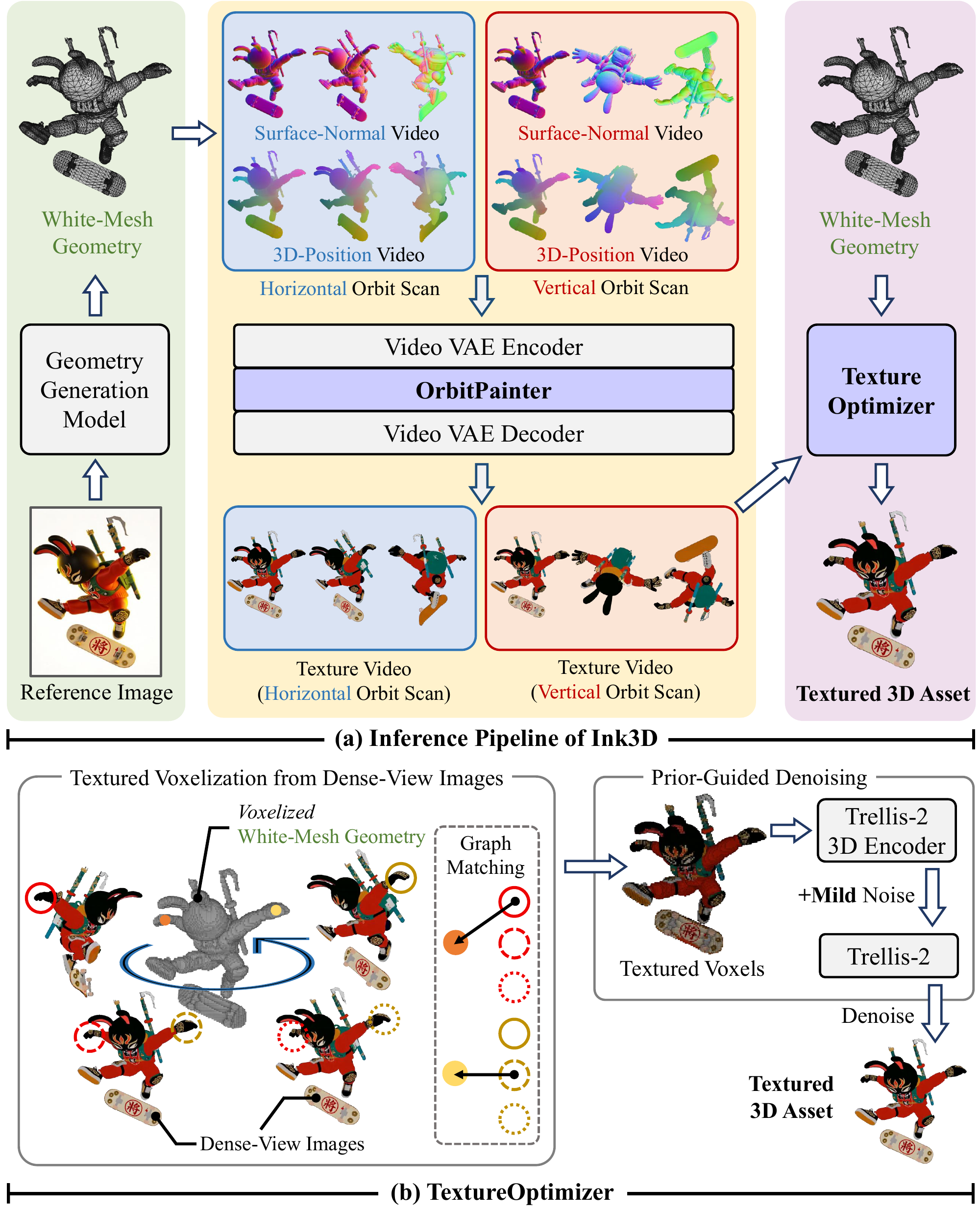}
    \vspace{-6mm}
    \caption{(a) \textbf{Inference pipeline of Ink3D.} Given a reference image, we first estimate the corresponding white-mesh geometry using an off-the-shelf geometry generation model. The trained OrbitPainter then generates orbit-scan videos of the object, whose frames are collected to form a dense-view image set. Finally, TextureOptimizer bakes these dense views onto the white-mesh geometry to produce the final textured 3D asset. (b) \textbf{TextureOptimizer.} We first convert the dense-view images into textured voxels based on the white-mesh geometry by formulating voxel–pixel correspondence as a graph-matching problem. The resulting textured voxels are then refined via prior-guided denoising: mild noise is added and Trellis-2 performs a few denoising steps to produce the final textured 3D asset.}
    \vspace{-4mm}
    \label{fig:inference}
\end{figure*}

\vspace{-1mm}
\subsection{TextureOptimizer}
\label{sec:TextureOptimizer}
\vspace{-1mm}
Once OrbitPainter is trained, we use it to generate dense multi-view texture observations for a given 3D geometry. Specifically, given a white-mesh geometry produced by an off-the-shelf geometry estimation model and a reference image that depicts the object's rough appearance, we first render surface-normal and 3D-position videos of the white mesh using the same orbit-scan trajectories described in Section~\ref{sec:OrbitPainter}.

These videos, together with the reference image, are encoded into latent representations using the video VAE encoder. The resulting latents, along with randomly sampled noise, are then fed into the OrbitPainter model. Through the iterative denoising process, OrbitPainter predicts texture latent videos, which are subsequently decoded by the video VAE decoder to produce orbit-scan texture videos for both horizontal and vertical scans, as illustrated in Figure~\ref{fig:inference}(a).

The generated orbit-scan texture videos provide a dense set of multi-view observations, where each frame corresponds to a single viewpoint. We denote this set as
$\mathcal{O}=\{\boldsymbol{O}_k\}_{k=1}^{K}$, where $K$ is the total number of views. Since each orbit scan produces 61 frames and we perform both horizontal and vertical scans, the resulting set contains $K=122$ views in total.

\noindent\textbf{Can Simple Baking Work?}
A straightforward solution is to directly use the dense view set $\mathcal{O}$ to bake textures onto the white-mesh geometry. However, in practice this naive approach produces noticeable artifacts on the resulting textured mesh, such as color bleeding and blurry textures. The root cause is that frames in the generated orbit-scan videos are not strictly geometry-consistent. Although the generated videos visually resemble an orbit scan around a 3D object, the same surface region may appear slightly inconsistent across different frames. Consequently, when these frames are reprojected back onto the mesh using the corresponding camera parameters, pixels that should correspond to the same 3D surface location may not align perfectly. This misalignment leads to inconsistent color accumulation during baking, resulting in texture artifacts.

\noindent\textbf{Motivation.}
The above observations inspire us to design a more robust baking mechanism that can mitigate the geometry inconsistency across dense views. We thus introduce \textit{TextureOptimizer}, a neural baking approach that leverages the capabilities of the native 3D generation model Trellis-2~\cite{xiang2025trellis2} in a training-free manner. The original Trellis-2 model generates a textured 3D asset from a geometry initialized with Gaussian noise (referred to as \textit{noisy geometry}), effectively transforming a fully noisy geometry into a textured 3D asset. However, while effective, Trellis-2 alone struggles to reproduce extremely complex textures. As illustrated in Figure~\ref{fig:inference}(b), the key idea of TextureOptimizer is simple: instead of starting from a fully noisy geometry, we initialize the geometry with \textit{mild} noise while incorporating as much texture information as possible from the dense-view image set $\mathcal{O}$. This initialization guides the model during the denoising process toward generating a textured 3D asset that faithfully follows the appearance observed in $\mathcal{O}$, while significantly reducing baking artifacts.

\noindent\textbf{Textured Voxelization from Dense-View Images.} We first convert the dense-view image set $\mathcal{O}$ into a \textit{textured voxel representation} based on the white-mesh geometry $\mathcal{G}$ to satisfy the input requirements of Trellis-2.

Concretely, given a set of occupied surface voxels 
$\mathcal{V}=\{\boldsymbol{v}_m\}_{m=1}^{M}$ containing $M$ voxels (obtained by voxelizing the white-mesh geometry $\mathcal{G}$), and a dense-view image set 
$\mathcal{O}=\{\boldsymbol{O}_k \in \mathbb{R}^{H\times W}\}_{k=1}^{K}$ consisting of $K$ multi-view images, each with spatial resolution $(H, W)$, our goal is to establish a globally optimal correspondence between the $M$ voxels and the $H \times W \times K$ image pixels. In other words, each voxel is assigned the color of one pixel selected from the dense-view image set. This correspondence should minimize texture seams while respecting geometric visibility constraints. 

To this end, we formulate the problem as a \textit{graph-matching} task and optimize the following objective using a Markov Random Field (MRF) over the 6-connected voxel graph $\mathcal{E}$:
\begin{equation}
\label{eq:mrf_energy}
\min_{\{k_m\}_{m=1}^M} \sum_{m=1}^{M} D_m(k_m) + \lambda \sum_{(m,n) \in \mathcal{E}} w_{m,n} \mathbf{1}[k_m \neq k_n],
\end{equation}
where $k_m \in \{1, \dots, K\}$ is the assigned view index for voxel $\boldsymbol{v}_m$. By determining $k_m$, voxel $\boldsymbol{v}_m$ is explicitly assigned the pixel value $\boldsymbol{O}_{k_m}(i_m, j_m)$, where $(i_m, j_m)$ represents the 2D projected coordinates of the voxel in the $k_m$-th view. The data term (the first term) $D_m(k_m)$ evaluates the geometric visibility and projection reliability of this specific pixel, which is measured based on the angle between the voxel's surface normal and the camera's viewing direction. The smoothness term (the second term) utilizes color-gradient-adaptive weights $w_{m,n}$ to penalize texture seams, which occur when adjacent voxels draw colors from different views (\ie, $k_m \neq k_n$). Finally, we optimize Eq.~\eqref{eq:mrf_energy} via alpha-expansion~\cite{boykov2001fast} and fill any unobserved voxels via breadth-first color diffusion. After optimization, we obtain a textured voxel representation of the geometry, denoted as $\mathcal{V}'$.



\noindent\textbf{Prior-Guided Denoising.} The original Trellis-2 model takes untextured voxels corrupted with randomly sampled Gaussian noise as input and gradually denoises them to produce textured geometry. As the denoising process progresses, the intermediate representations gradually accumulate texture information.

In our setting, however, we already obtain textured voxels $\mathcal{V}'$ that contain substantial texture information. This observation motivates us to leverage Trellis-2 to perform a \textit{prior-guided denoising} process. Specifically, as illustrated in Figure~\ref{fig:inference}(b). we first use the Trellis-2 3D VAE encoder to extract latent representations from $\mathcal{V}'$. We then perturb the latents with \textit{mild} Gaussian noise and feed the resulting mildly noisy latents into Trellis-2. Since the initialization is only mildly corrupted, the model performs only a few denoising steps (corresponding to the later stages of the original denoising trajectory) to generate the final textured geometry.

\vspace{-3mm}
\section{Experiment}
\vspace{-1mm}

\noindent\textbf{Implementation Details.} \emph{OrbitPainter} is trained on a subset of the Objaverse~\cite{deitke2023objaverse} and sketchfab~\cite{sketchfab2025}  dataset, consisting of 60K samples with albedo textures. The model is initialized from Wan-2.2-14B~\cite{wan2025wan}. The geometry-aware spatial attention module is initialized using the weights of the corresponding full-attention layers. During training, we adopt LoRA-style fine-tuning with a rank of 32. Optimization is performed using the AdamW~\cite{adamw} optimizer with a learning rate of 1e-4. The model is trained for 45k iterations with a batch size of 1 on 8$\times$ NVIDIA A100 GPUs. Unless otherwise specified, we generate two orbit-scan videos for each object: one with horizontal camera rotation and one with vertical rotation. Each video has a resolution of $512\times512$ and contains 61 frames. Additional implementation details are provided in the appendix.

\noindent\textbf{Evaluation Data.} We select 85 high-quality samples from the Texverse~\cite{zhang2025texverse} dataset. As our focus is on generating highly complex textures, the original assets in Texverse do not sufficiently meet this requirement. Therefore, for each sample we retain only its geometry and discard the original texture. We then render the frontal-view projection of the white mesh and feed it into the commercial image generation model Nano Banana to synthesize a reference image with complex textures. As a result, for each of the 85 samples we obtain a pair consisting of a white-mesh geometry and a corresponding reference image that exhibits rich and complex textures.

\noindent\textbf{Evaluation Metrics.} For each of the 85 samples, we provide its ground-truth geometry and the reference texture image to our method as well as the compared methods to generate textured 3D assets. We then render the front-view projection of each generated asset and compute several image-level metrics against the reference image. Specifically, we report \textit{FID}, \textit{CLIP-FID}, \textit{LPIPS}, and \textit{CLIP-I}. FID measures the Fréchet distance between the feature distributions of generated images and reference images in the Inception feature space, reflecting overall visual quality and realism. CLIP-FID replaces the Inception features with CLIP features, providing a metric that better captures semantic alignment with the reference. LPIPS measures perceptual similarity between two images based on deep features, reflecting low-level visual fidelity. Finally, CLIP-I computes the cosine similarity between CLIP image embeddings of the generated image and the reference image, evaluating semantic consistency between them.

\begin{table}[!t]
    \centering
    \caption{Quantitative comparison of texture synthesis quality. Our primary competitor is Trellis-2, a recently released state-of-the-art 3D generation model. Ink3D significantly outperforms Trellis-2 and other existing open-source approaches in texture synthesis.}
    \vspace{-3mm}
    \begin{tabular}{lccccc}
    \toprule
    Method &  FID$\downarrow$ &  CLIP-FID$\downarrow$ &  LPIPS$\downarrow$ & CLIP-I$\uparrow$ \\
    \midrule
    Paint3D~\cite{zeng2024paint3d} & 146.0  & 570.6 & 0.3417  &  0.7980  &   \\
    TEXGEN~\cite{yu2024texgen} & 123.7  & 291.8  & 0.2280  & 0.8409  &   \\
    SeqTex~\cite{yuan2025seqtex} & 148.3 & 387.9 & 0.2601 & 0.8052  \\
    Trellis-2~\cite{xiang2025trellis2} & 120.4 & 223.9 & 0.2714 & 0.8694  \\
    \midrule
    Ink3D (Ours) & 103.7 &  195.6 & 0.2029 & 0.8979  \\
    \bottomrule
    \end{tabular}
    \vspace{-3mm}
    \label{tab:main}
\end{table}

\vspace{-2mm}
\subsection{Main Results}
\vspace{-1mm}
Table~\ref{tab:main} presents a quantitative comparison between Ink3D and several representative open-source approaches capable of generating textures given a white-mesh geometry. Our primary comparison is with Trellis-2, a recently released state-of-the-art 3D generation model that supports both joint geometry–texture generation and texture synthesis conditioned on a given geometry and reference image. To ensure a fair comparison, we adopt the second mode of Trellis-2, where the model is provided with the same white-mesh geometry and reference image as inputs. As shown in Table~\ref{tab:main}, Ink3D significantly outperforms Trellis-2 and other competing methods in texture synthesis quality. This improvement highlights the advantage of introducing geometry-aware video generation into the texture creation pipeline, enabling Ink3D to synthesize richer and more detailed surface appearances. Additional qualitative results are provided in the supplementary material.

\vspace{-2mm}
\subsection{Ablation Studies} 
\vspace{-3mm}

\begin{table}[!t]
    \centering
    \caption{Ablation study on leveraging 3D geometry priors in OrbitPainter. OrbitPainter incorporates geometry priors to generate geometry-consistent orbit-scan videos at two levels: (1) \textit{input level}, by directly providing a 3D-position video and a surface-normal video as additional inputs; and (2) \textit{network level}, by introducing the proposed Geometry-Aware Sparse Attention (GASA), which constrains attention to be performed among pixels corresponding to the same 3D location across frames.}
    \vspace{-3mm}
    \begin{tabular}{lccccc}
    \toprule
    Setting &  FID$\downarrow$ &  CLIP-FID$\downarrow$ & LPIPS$\downarrow$ & CLIP-I$\uparrow$ \\
    \midrule
    Baseline & 103.7 & 195.6 & 0.2030 & 0.8980\\
    w/o Surface Normal & 113.9 & 218.3 & 0.2128 & 0.8841\\
    w/o 3D-Position & 149.2 & 361.5 & 0.2367 & 0.8282\\
    w/o GASA & 109.9 & 203.7 & 0.2077 & 0.8893\\
    \bottomrule
    \end{tabular}
    \vspace{-1mm}
    \label{tab:component}
\end{table}

\noindent\textbf{3D Geometry Prior in OrbitPainter.} As introduced in Section~\ref{sec:OrbitPainter}, OrbitPainter is trained with two conditioning priors: (1) a 2D texture appearance prior provided by the reference image, and (2) a 3D geometry prior provided by the white-mesh geometry. The latter prior goes beyond traditional image-to-video generation settings. To incorporate this 3D geometry prior, we introduce two additional conditioning signals: a surface-normal video and a 3D-position video. In Table~\ref{tab:component}, we compare our default model with two ablated variants: one without the surface-normal video prior and another without the 3D-position video prior. As shown in Table~\ref{tab:component}, removing either prior degrades the performance, with the removal of the 3D-position prior causing a particularly significant drop. This is because the 3D-position signal provides strong geometric guidance, helping OrbitPainter maintain geometry-consistent appearance across orbit-scan frames.

\noindent\textbf{Geometry-Aware Sparse Attention.}
In Section~\ref{sec:sparseattention}, we introduce a Geometry-Aware Sparse Attention (GASA) layer, which operates in parallel with the full attention layer, as illustrated in Figure~\ref{fig:training}(b). GASA explicitly encourages the network to learn correspondences across frames by linking pixels that share the same original 3D location when re-projected into 3D space. In Table~\ref{tab:component}, we compare our default model with a variant that removes GASA. The observed performance drop demonstrates that better utilization of geometric priors improves geometry consistency across the generated orbit-scan frames, which in turn leads to higher-quality textured 3D assets. While the injected 3D-position and surface-normal videos introduce geometry priors at the input level, GASA incorporates geometry priors directly within the network attention mechanism.

\begin{table}[!t]
    \centering
    \caption{Ablation study on the impact of orbit-scan video resolution and frame count on geometry texturing performance.}
    \vspace{-3mm}
    \begin{tabular}{ccccccc}
    \toprule
    Resolution & \#Frame &  FID$\downarrow$ &  CLIP-FID$\downarrow$ & LPIPS$\downarrow$ & CLIP-I$\uparrow$ \\
    \midrule
    384$^2$ & 61 & 124.2 & 269.4 & 0.2221 & 0.8694\\
    512$^2$ & 61 & 103.7 & 195.6 & 0.2030 & 0.8980\\
    512$^2$ & 41 & 122.8 & 215.3 & 0.1909 & 0.8988\\
    512$^2$ & 25 & 124.8 & 215.2 & 0.1958 & 0.8963\\
    \bottomrule
    \end{tabular}
    \vspace{-4mm}
    \label{tab:resolution-frame}
\end{table}

\noindent\textbf{Video Quality Determines Texture Quality.} We study two factors that affect the quality of the orbit-scan videos generated by OrbitPainter: the video resolution and the number of frames. The first directly influences the fidelity of the final baked texture, as higher-resolution frames provide richer visual details. The second controls the effective camera motion speed. Since the generated video always corresponds to a full 360-degree orbit scan, using fewer frames results in faster camera motion. Table~\ref{tab:resolution-frame} analyzes the impact of these two factors. The results suggest that, when computational cost is not the primary concern, higher-resolution video generation leads to better final texture quality. In addition, using too few frames significantly degrades performance. This is mainly because faster camera motion introduces motion blur in the generated orbit-scan videos, which subsequently leads to blurred texture artifacts after baking.

\begin{table}[!t]
    \centering
    \caption{Comparison of our TextureOptimizer baking strategy with traditional methods, namely Average Baking and Differentiable Rendering, for baking dense-view image textures onto the geometry.}
    \vspace{-3mm}
    \begin{tabular}{lccccc}
    \toprule
    Baking Strategy &  FID$\downarrow$ &  CLIP-FID$\downarrow$ &  LPIPS$\downarrow$ & CLIP-I$\uparrow$ \\
    \midrule
    Average Baking & 121.0 & 227.7 & 0.2498 & 0.8888\\
    Differentiable Rendering & 123.3 & 226.8 & 0.2746 & 0.8944\\
    \midrule
    TextureOptimizer & 103.7 & 195.6 & 0.2030 & 0.8980\\
    \bottomrule
    \end{tabular}
    \vspace{-4mm}
    \label{tab:bake}
\end{table}

\noindent\textbf{TextureOptimizer.}
In Section~\ref{sec:TextureOptimizer}, we describe how to bake the dense-view image set (containing all orbit-scan frames) onto the white-mesh geometry, and introduce \emph{TextureOptimizer}, a training-free neural baking strategy. In Table~\ref{tab:bake}, we compare TextureOptimizer with two baseline baking methods: (1) \textit{Average Baking}: each pixel from every view is re-projected onto the geometry, and the final texture value at each 3D location is obtained by averaging all pixels projected to that location.
(2) \textit{Differentiable Rendering}: widely used in modern 3D generation models such as Trellis-1~\cite{xiang2025trellis1}, where a texture map is optimized through gradient descent so that rendering the textured 3D model matches the dense-view observations as closely as possible. Since the generated orbit-scan videos are not perfectly geometry-consistent, all approaches must handle this issue. The results in Table~\ref{tab:bake} suggest that TextureOptimizer better integrates dense-view observations while reducing artifacts caused by cross-view inconsistencies.
 
\vspace{-3mm}
\section{Conclusion}
\vspace{-2mm}
We introduce Ink3D, a framework that bridges 3D generation and large-scale video generative models to synthesize extremely complex textures for 3D assets. Given a reference image, Ink3D first reconstructs a white-mesh geometry using an off-the-shelf 3D generation model, then employs a conditional video model, OrbitPainter, to generate dense 360° orbit-scan videos for comprehensive multi-view observations. These observations are integrated by TextureOptimizer, a neural baking framework that consolidates multi-view appearance while mitigating artifacts caused by geometric inconsistencies. Leveraging the strong visual modeling capability of large-scale video models, Ink3D produces rich textures that are difficult for existing 3D generation methods to achieve.

\section*{Acknowledgements}
This work is supported by the Zhongguancun Academy, (Grant No.s 20240307).

\clearpage 
\begingroup \centering {\Large\bfseries Supplementary Material\par} \endgroup
\setcounter{section}{0} \setcounter{figure}{0} \setcounter{table}{0} \setcounter{equation}{0} 
\renewcommand{\thesection}{\Alph{section}} 
\renewcommand{\thefigure}{A\arabic{figure}} \renewcommand{\thetable}{A\arabic{table}} \renewcommand{\theequation}{A\arabic{equation}}

\vspace{2em}  





\appendix

\noindent We provide supplementary material for a deeper understanding and more analysis related to the main paper, arranged as follows:
\begin{enumerate}
    \item Qualitative Comparison (Appendix~\ref{sup:additional_qualitative_results})
    \item Variety Test (Appendix~\ref{sup:variety})
    \item Ablations (Appendix~\ref{sup:ablation})
    \item Inference Cost/ Latency (Appendix~\ref{sup:cost})
    \item Limitations and Future Work (Appendix~\ref{sup:limitations})
    
\end{enumerate}


\section{Qualitative Comparison}
\label{sup:additional_qualitative_results}
 
\begin{figure*}[!p]
  \centering
    \includegraphics[width=0.99\linewidth]{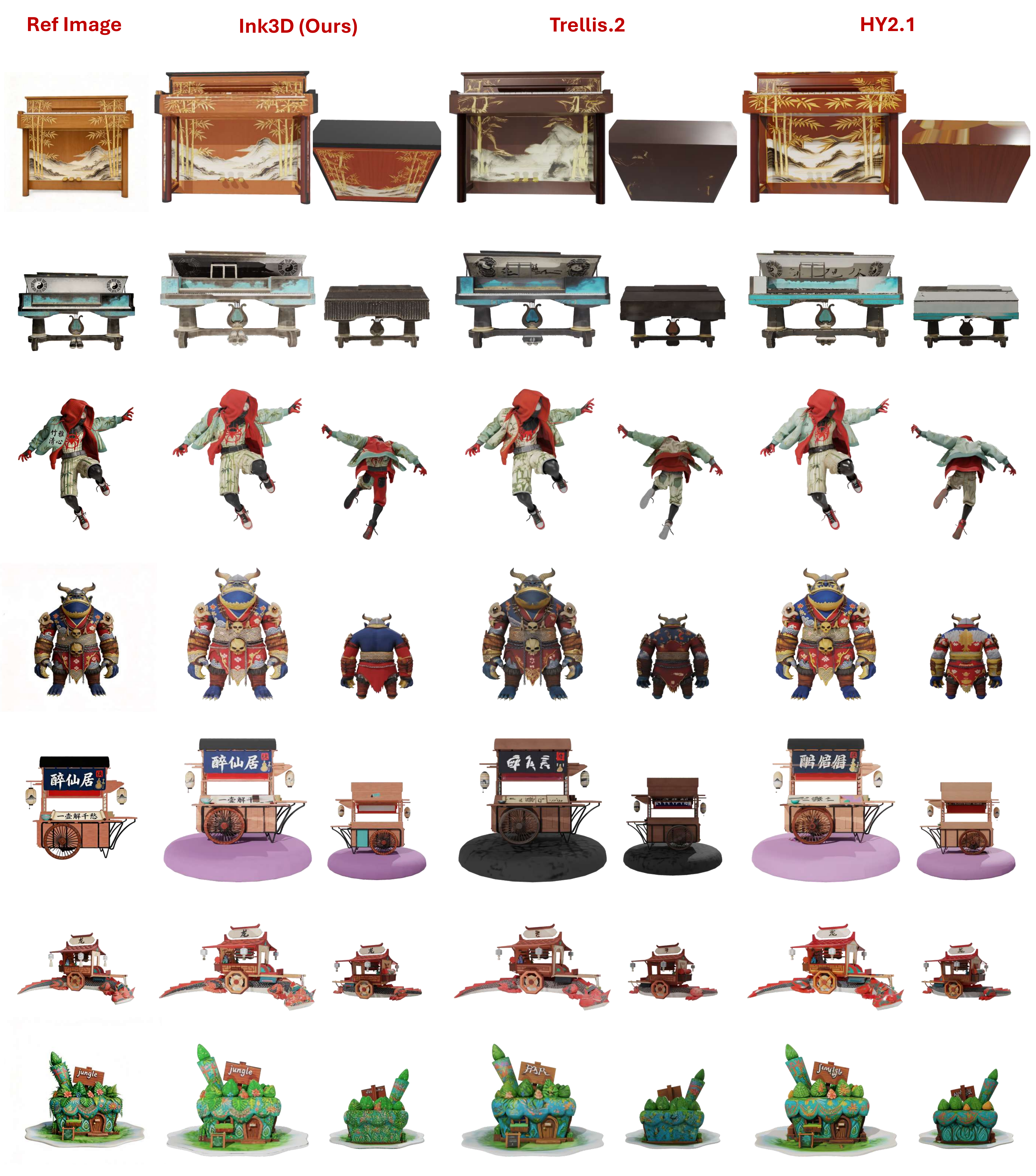}
    \caption{\textbf{Qualitative Comparison.}}
    \label{fig:qualitative_comparison}
\end{figure*}

\begin{figure*}[!p]
  \centering
    \includegraphics[width=0.99\linewidth]{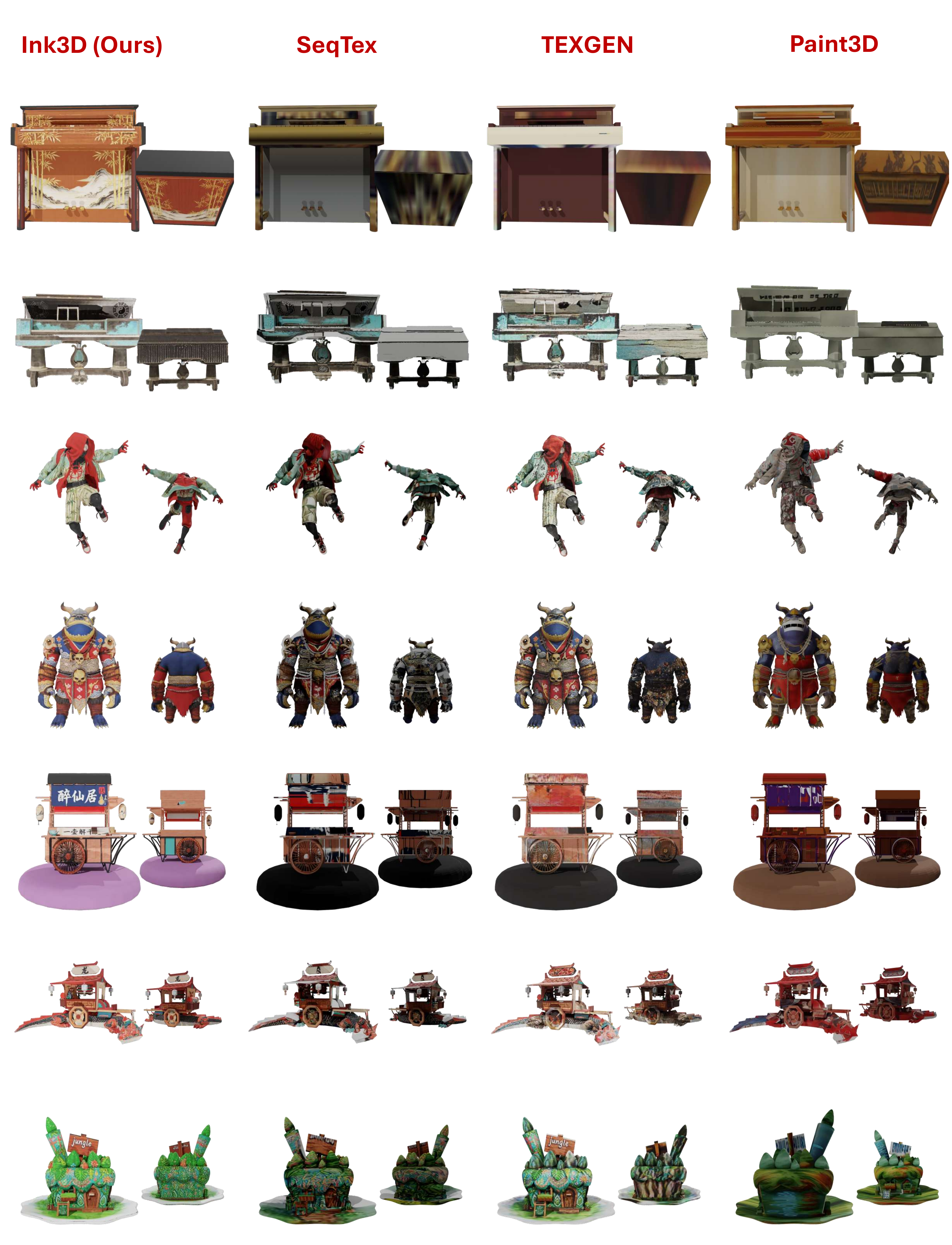}
    \caption{\textbf{Qualitative Comparison.}}
    \label{fig:qualitative_comparison2}
\end{figure*}

Fig.~\ref{fig:qualitative_comparison} and Fig.~\ref{fig:qualitative_comparison2} presents a qualitative comparison with recent SOTA methods, including TRELLIS.2~\cite{xiang2025trellis2}, HY2.1~\cite{hunyuan3d2025}, SeqTex~\cite{yuan2025seqtex}, TEXGEN~\cite{yu2024texgen}, and Paint3D~\cite{zeng2024paint3d}.  
Among these, TRELLIS.2 is a native 3D approach; however, due to the scarcity of high-quality 3D texture data, it struggles to render complex textures on flat geometries.
HY2.1, SeqTex, and Paint3D, like our method, follow a multi-view generate-then-bake paradigm. 
Nevertheless, our approach achieves superior fidelity in high-frequency texture details—especially for text. 
This is attributed to three key factors: 
\textit{\textbf{(1)}} maximal preservation of WAN's powerful generative capabilities, 
\textit{\textbf{(2)}} frame-to-frame smoothness ensured by video generation, and
\textit{\textbf{(3)}} an optimization-based baking strategy applied directly in native 3D space.
Additionally, both SeqTex and TEXGEN opt to generate textures directly in UV space instead of baking. 
This imposes stringent requirements on the viewpoints of reference images. 
The references must be well-aligned in viewpoint to enable reliable back-projection into UV space for subsequent completion. 
Otherwise, it is likely to produce unsatisfactory results. 
While SeqTex markedly reduces the pressure of texture completion compared to TEXGEN by leveraging video generation, it faces limitations because such models are ill-suited for generating fast-spinning sequences of orthogonal views. Consequently, this results in compromised detail fidelity and consistency across the generated multi-views.

\section{Variety Test}
\label{sup:variety}
\begin{figure*}[!p]
  \centering
    \includegraphics[width=0.99\linewidth]{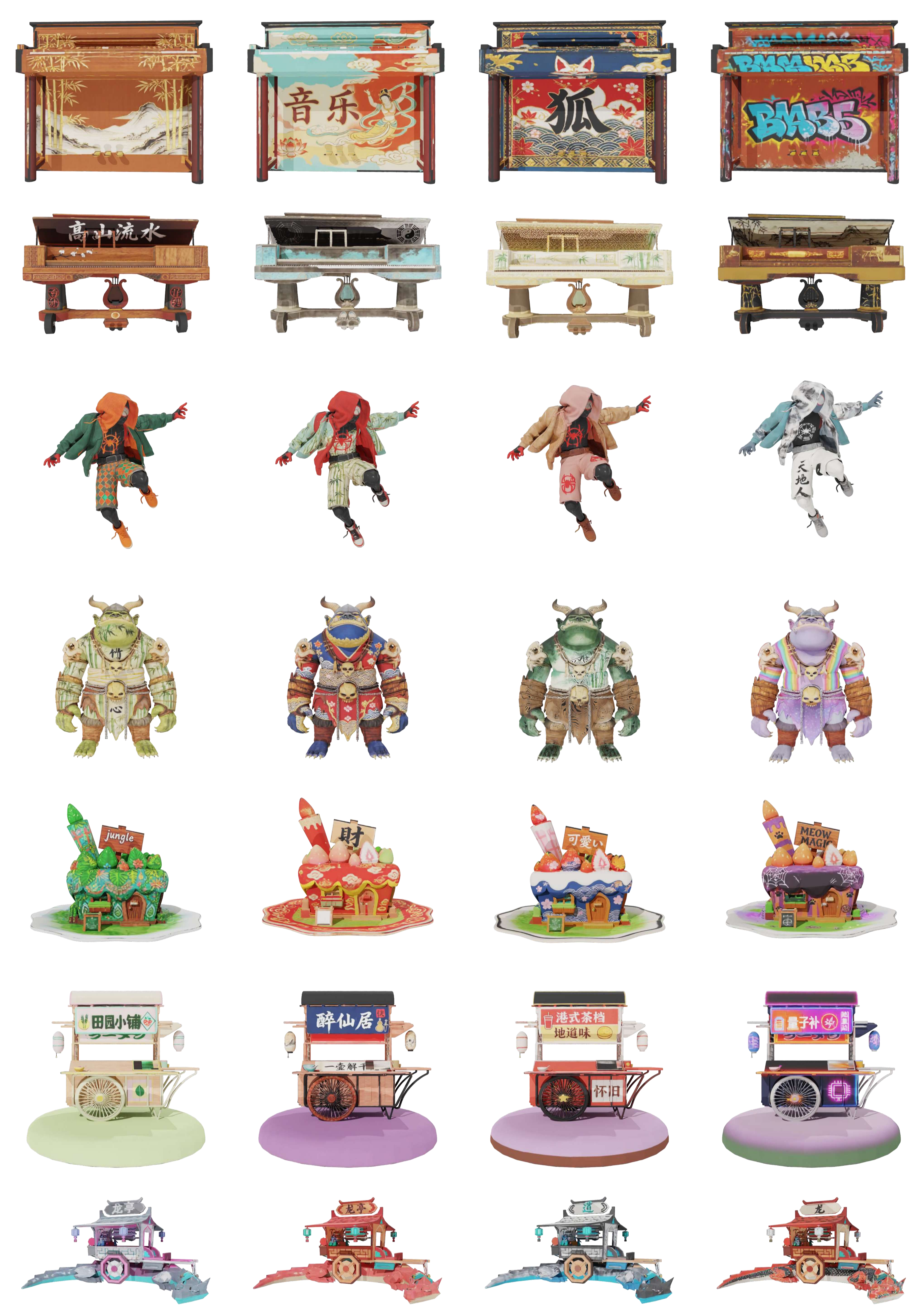}
    \caption{\textbf{Variety Test.}}
    \label{fig:variety_test}
\end{figure*}

Fig.~\ref{fig:variety_test} presents additional generation results. Leveraging the powerful capabilities of WAN~\cite{wan2025wan}, Ink3D demonstrates strong generalization across a diverse range of styles. Furthermore, thanks to its 3D-space baking strategy, Ink3D effectively preserves complex texture details, achieving high fidelity even for text, which is particularly sensitive to distortion.

\section{Ablations}
\label{sup:ablation}
\subsection{Ablations on Video Generation}

\begin{figure*}[!th]
  \centering
    \includegraphics[width=0.99\linewidth]{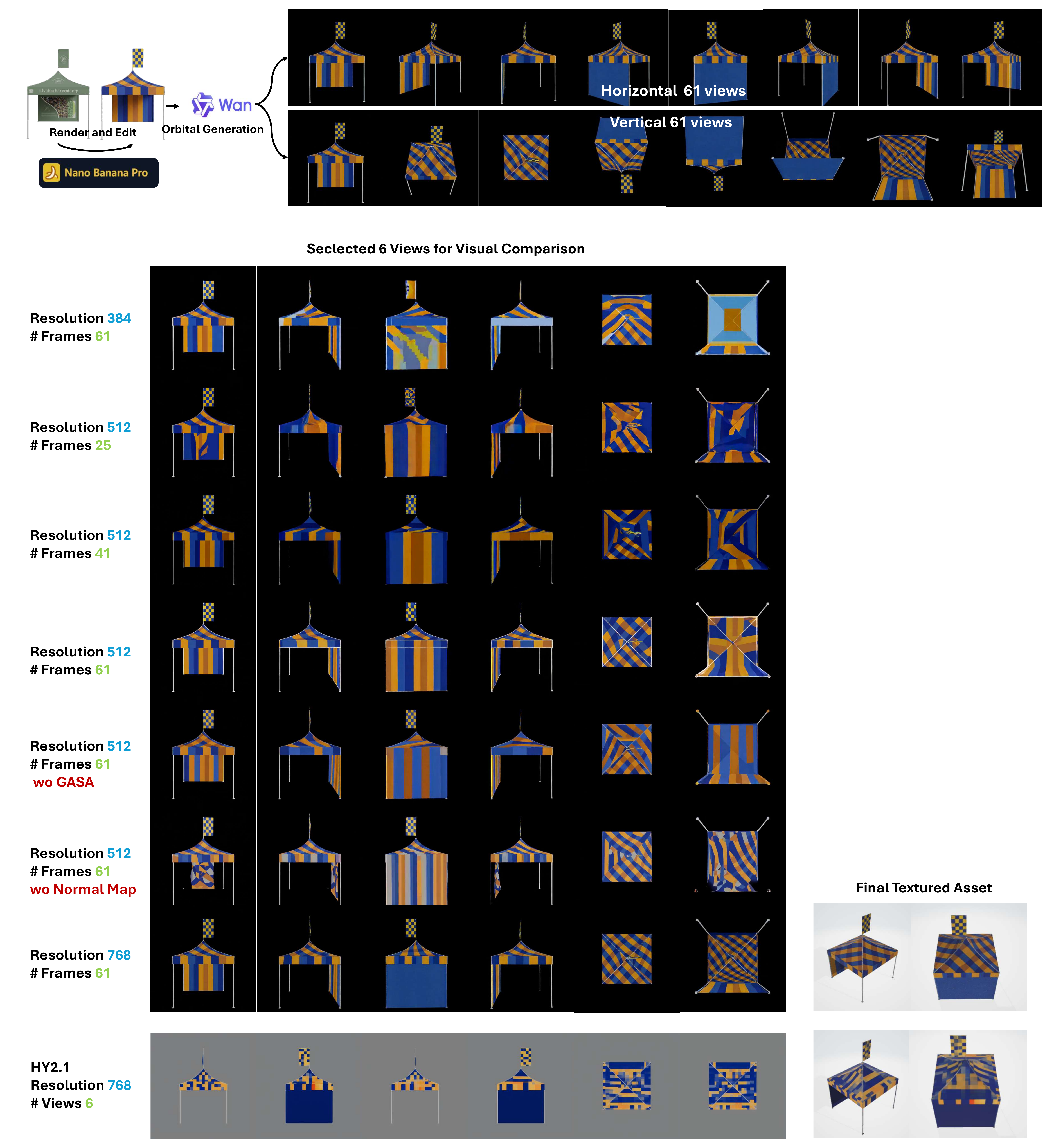}
    \caption{\textbf{Ablations on Video Generation.}}
    \label{fig:video_abla}
\end{figure*}

For clearer evaluation of multi-view consistency, we use a flat-geometry asset (tent) edited via nanobanana to generate a grid-textured reference image.
In Fig.~\ref{fig:video_abla}, we find that reduced resolution and fewer input views (\ie, faster rotation) drastically diminish multi-view consistency.
Therefore, \textbf{Ink3D aims to offer a new perspective: }
while prior methods have leveraged video generation models to improve multi-view consistency, they suffer from insufficient view density. 
We argue that \textbf{video models trained on massive real-world data inherently possess robust 3D consistency.}
Consequently, when the generation domain aligns closely with the data domain, it is possible to achieve 3D-consistent textures with high-fidelity details.
Despite the largely increased computational cost, we strive to demonstrate the \textbf{upper bound} of video generation models in preserving complex texture details.

\clearpage
\begin{figure*}[p]
  \centering
  \includegraphics[width=0.99\textwidth]{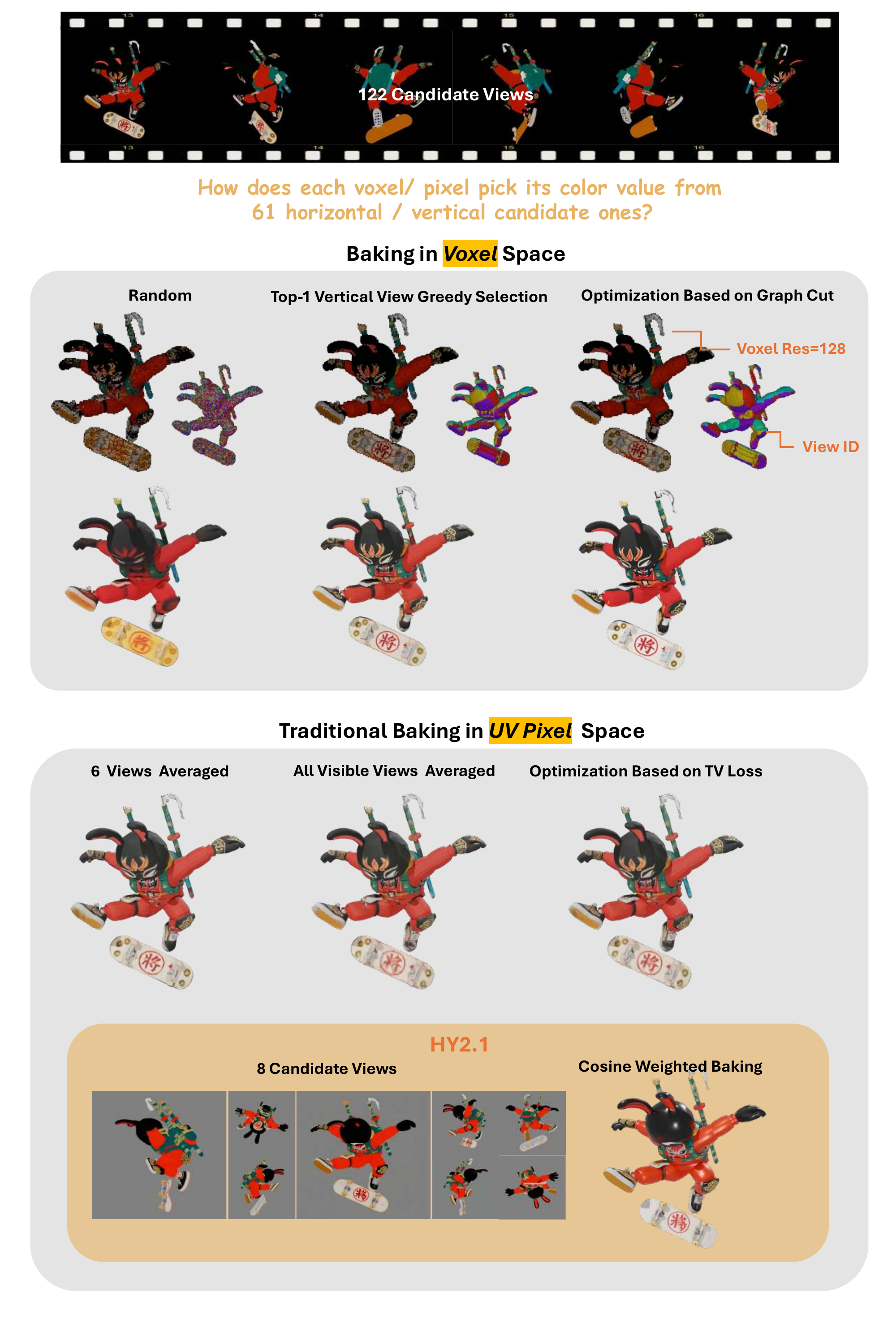}
  \caption{\textbf{Ablations on Baking.}}
  \label{fig:bake_abla}
\end{figure*}
\clearpage

\subsection{Ablations on Baking}

Fig.~\ref{fig:bake_abla} illustrates that while video generation provides dense, smooth, and reliable viewpoints, previous baking methods encounter two critical issues when voxels are presented with an abundance of candidate color values:
\textit{\textbf{(1)}} Texture Blurring on Flat Geometries: Due to the inherent averaging across all viewpoints, increasing the number of views leads to blurrier texture results on flat surfaces.
\textit{\textbf{(2)}} Color Bleeding on Complex Geometries: Color bleeding artifacts caused by view misalignment at complex geometric boundaries remain obvious.
To address the blurring of high-frequency textures, we propose a GraphCut-based optimization algorithm that effectively mitigates this issue. 
Regarding color bleeding at complex boundaries, we leverage TRELLIS.2's native 3D perception of intricate geometries. 
Specifically, we convert the model into an O-voxel representation and project multi-view colors directly into this voxel space for baking and following refinement, thereby preserving sharp boundary details.




\section{Inference Cost/ Latency}
\label{sup:cost}
On a single H100 GPU, OrbitPainter generates two 61-frame $512^2$ orbit videos in \textit{$\sim$12 minutes} with \textit{45G} GPU memory, while TextureOptimizer takes \textit{$\sim$1 minute}, resulting in a total cost of \textit{$\sim$13 minutes}. The video generation dominates the overall latency. Using DMD2 acceleration can shorten video generation \textit{from $\sim$12 minutes to $\sim$2 minutes}. 

\section{Limitations and Future Work}
\label{sup:limitations}

Like most approaches that rely on video generation models, our method inherits the high inference latency associated with generating high-resolution videos. Although Ink3D can synthesize highly complex textures, the video generation stage remains computationally expensive. Future work may explore acceleration techniques, such as DMD2~\cite{yin2024improved}, to reduce inference cost.

%
%
\bibliographystyle{splncs04}
\bibliography{main}
\end{document}